# Globally Variance-Constrained Sparse Representation and Its Application in Image Set Coding

Xiang Zhang, Jiarui Sun, Siwei Ma, *Member, IEEE*, Zhouchen Lin, *Fellow, IEEE*, Jian Zhang, *Member, IEEE*, Shiqi Wang, *Member, IEEE*, and Wen Gao, *Fellow, IEEE*

*Abstract*—Sparse representation leads to an efficient way to approximately recover a signal by the linear composition of a few bases from a learnt dictionary based on which various successful applications have been achieved. However, in the scenario of data compression, its efficiency and popularity are hindered. It is because of the fact that encoding sparsely distributed coefficients may consume more bits for representing the index of nonzero coefficients. Therefore, introducing an accurate rate constraint in sparse coding and dictionary learning becomes meaningful, which has not been fully exploited in the context of sparse representation. According to the *Shannon* entropy inequality, the variance of Gaussian distributed data bound its entropy, indicating the actual bitrate can be well estimated by its variance. Hence, a globally variance-constrained sparse representation (GVCSR) model is proposed in this paper, where a variance-constrained rate term is introduced to the optimization process. Specifically, we employ the alternating direction method of multipliers (ADMMs) to solve the non-convex optimization problem for sparse coding and dictionary learning, both of them have shown the state-of-the-art rate-distortion performance for image representation. Furthermore, we investigate the potential of applying the GVCSR algorithm in the practical image set compression, where the optimized dictionary is trained to efficiently represent the images captured in similar scenarios by implicitly utilizing inter-image correlations. Experimental results have demonstrated superior rate-distortion performance against the state-of-the-art methods.

## I. INTRODUCTION

LOSSY image compression aims to reduce redundancies among pixels while maintaining the required quality for the purpose of efficient transmission and storage. Due to the energy compaction property, transform coding approaches have shown great power in de-correlation and have been widely adopted in image compression standards, as witnessed from the Discrete Cosine Transform (DCT) in Joint Photographic Experts Group (JPEG) [1] to the Discrete Wavelet Transform (DWT) in JPEG2000 [2]. The common property of DCT and DWT lies in that their basis functions are orthogonal to each other and meanwhile fixed despite the characteristics of the input signals. However, such inflexibility may greatly restrict their representation efficiency.

In the early 1990s, Olshausen and Field proposed the sparse coding with a learnt dictionary [3] to represent an image. Since then the sparse theory has been widely studied and advocated [4], [5]. It is widely believed that the sparsity property is efficient in dealing with the rich, varied and directional information contained in natural scenes [4], [6]. It has also been proven that the basis in sparse coding has the characteristics of spatially localized, oriented and bandpass, which are closely relevant to the properties of the receptive fields of simple cells. Recent studies further validated the idea that the sparse coding performs in a perceptual way that mimics the Human Visual System (HVS) on natural images [7]–[9]. Based on the sparse model, numerous applications have been successfully achieved, including image denoising [10], restoration [11]–[14], quality assessment [15]–[17], etc.

Towards image compression, the sparse coding need to be optimized in terms of both rate and distortion [18]. Despite the fact that the sparse coding can provide more efficient representation than orthogonal transforms [19], its efficiency in compression tasks is however limited. In the conventional sparse representation model, the objective function consists of two parts, the data fidelity term and the sparsity constraint term.

Manuscript received December 12, 2016; revised October 2, 2017 and January 19, 2018; accepted March 22, 2018. Date of publication April 5, 2018; date of current version April 26, 2018. This work was supported in part by the National Natural Science Foundation of China under Grant 61571017 and in part by the National Postdoctoral Program for Innovative Talents under Grant BX201600006. The work of Z. Lin was supported in part by the National Basic Research Program of China (973 Program) under Grant 2015CB352502, in part by the National Natural Science Foundation of China under Grant 61625301 and Grant 61731018, and in part by Qualcomm and Microsoft Research Asia. The associate editor coordinating the review of this manuscript and approving it for publication was Dr. Catarina Brites. *(Corresponding author: Siwei Ma.)*

X. Zhang, S. Ma, and W. Gao are with the Institute of Digital Media, School of Electronics Engineering and Computer Science, Peking University, Beijing 100871, China (e-mail: x_zhang@pku.edu.cn; swma@pku.edu.cn; wgao@pku.edu.cn).

J. Sun is with Tencent Computer System Co. Ltd., Shenzhen 518057, China (e-mail: jiaruisun@tencent.com).

Z. Lin is with the Key Laboratory of Machine Perception (Ministry of Education), School of Electronics Engineering and Computer Science, Peking University, Beijing 100871, China, and also with the Cooperative Medianet Innovation Center, Shanghai Jiao Tong University, Shanghai 200240, China (e-mail: zlin@pku.edu.cn).

J. Zhang is with the School of Electronic and Computer Engineering, Peking University Shenzhen Graduate School, Shenzhen 518055, China (e-mail: zhangjian@pkusz.edu.cn).

S. Wang is with the Department of Computer Science, City University of Hong Kong, Kowloon, Hong Kong (e-mail: shiqwang@cityu.edu.hk).

Color versions of one or more of the figures in this paper are available online at http://ieeexplore.ieee.org.

Digital Object Identifier 10.1109/TIP.2018.2823546

The sparsity constraint is typically a $\ell_0$ regularization, i.e. the number of nonzero coefficients. Though more nonzero coefficients usually cost more coding bits, it may not be able to accurately reflect the actual coding bits. This motivates researchers to investigate advanced rate estimation algorithms for the rate-distortion optimized sparse representation. The Rate-Distortion Optimized Matching Pursuit (RDOMP) approaches [20]–[22] were proposed to address this issue, where the coding rate was estimated based on the probabilistic model of the coefficients. In [23], a shallow autoencoder algorithm was introduced, which can be viewed as an extension of sparse coding. The M-term pursuit [24] was presented for accelerating the matching pursuit process. Despite their performance improvements on image compression, they have some limitations. First, such matching pursuit based methods may suffer from the instability in obtaining the sparse coefficients [25]. Second, they operate and encode each sample separately, ignoring the data structure information and lacking global constraint over all input samples. This may cause two similar blocks having quite different representations and decrease the coding performance [26]–[28]. Third, the RDOMP methods can not be easily incorporated with the dictionary learning algorithm and such inconsistency may decrease the efficiency.

To address these weaknesses, the Globally Variance-Constrained Sparse Representation (GVCSR) model is proposed in this work, where a variance-constraint term is introduced into the objective function of sparse representation. By incorporating the rate model, minimizing the objective function turns out to be a joint rate-distortion optimization problem, which can be efficiently solved by Alternating Direction Method of Multipliers (ADMM) [29]. Moreover, it is solved in a "global" way, which significantly distinguishes from the separate manner of the matching pursuit. Therefore, such optimization based method can effectively utilize the intrinsic data structure and reduce the instability.

The major contributions of this work are as follows.
- We propose a novel sparse representation model, by which the rate-distortion jointly optimized sparse coding and dictionary learning can be coherently achieved.
- We solve the non-convex optimization problem with the ADMM. As such, the sparse coefficient correlation between similar image patches can be implicitly taken into account during optimization. It distinguishes from the patch-wise Matching Pursuit (MP) methods.
- We demonstrate the effectiveness and potential of the proposed method in the image set compression, where the dictionary is trained to represent similar images by implicitly reducing the inter-image redundancies. Experimental results have demonstrated better coding gains than other competitive methods.

The additional novelty of this paper over our previous work in [30] includes: 1) further improvements on the proposed method and more details are given; 2) More comparisons, analyses and discussions with the state-of-the-art matching pursuit and dictionary learning methods are provided; 3) The proposed method is further applied to image set compression.

For brevity, we summarize frequently used notations in Table I. The rest of the paper is organized as follows.

TABLE I
SUMMARY OF NOTATIONS

| Notation | Meaning |
| --- | --- |
| Bold capital letter | A matrix. |
| Bold lowercase letter | A vector. |
| $M^T$ | Transpose of matrix $M$. |
| $M_{i,j}$ | The $(i,j)^{th}$ element of matrix $M$. |
| $M_i$ | The $i^{th}$ column of matrix $M$. |
| $M^\dagger$ | Moore-Penrose pseudo-inverse of matrix $M$. |
| $\|\cdot\|_0$ | Number of nonzero elements. |
| $\|\cdot\|_1$ | $\|M\|_1 = \sum_{i,j} |M_{ij}|$. |
| $\|\cdot\|_F$ | Frobenious norm, $\|M\|_F = \sqrt{\sum_{i,j} M_{ij}^2}$. |
| $\|\cdot\|_2$ | Vector Euclidean norm, $\|v\|_2 = \sqrt{\sum_i v_i^2}$. |
| $tr(\cdot)$ | Sum of the diagonal entries of a matrix. |
| $\langle A, B \rangle$ | Inner product of matrix, $\langle A, B \rangle = tr(A^T B)$. |

The related works are briefly summarized in Section II. In Section III, we present the GVCSR model and the strategy to solve it via ADMM. Section IV evaluates the efficiency of GVCSR and further demonstrates its potential in image set compression. Section V concludes with a summary.

## II. RELATED WORKS

Sparse theory claims that signals can be well recovered by a few bases from an over-complete dictionary, where the dictionary is assumed to be highly adaptive to a set of signals within a limited subspace. The two basic problems in sparse representation are dictionary learning and sparse decomposition (a.k.a. sparse coding). In particular, the objective function of dictionary learning can be formulated as follows,

$$(D, \{A_i\}) = \arg\min_{D, \{A_i\}} \sum_i \|T_i - DA_i\|_2^2, \text{ s.t. } \|A_i\|_0 \leqslant L,$$
$$\|D_j\|_2^2 \leqslant 1, \quad \forall j \in \{1, 2, \ldots, M\} \quad (1)$$

where $D \in \mathbb{R}^{N \times M}$ is the redundant dictionary with $M$ bases. $T_i \in \mathbb{R}^{N \times 1}$ indicates the training data. $A_i \in \mathbb{R}^{M \times 1}$ is the corresponding sparse coefficients, whose $\ell_0$ norm is constrained by a given sparse level $L$. This is a non-convex optimization problem which is difficult to solve due to the $\ell_0$ norm. Typical algorithms for dictionary learning include the Method of Optimal Directions (MOD) [31], [32], K-SVD [33], Online Dictionary Learning (ODL) [34], Recursive Least Square (RLS) [35] and Sequential Generalization of K-means (SGK) [36].

With respect to the trained dictionary $D$, the sparse decomposition calculates the corresponding coefficients $A_i$ for the input signal $S_i$,

$$A_i = \arg\min_{A_i} \|S_i - DA_i\|_2^2, \quad \text{s.t. } \|A_i\|_0 \leqslant L, \quad (2)$$

which is a subproblem of (1). Several suboptimal solutions have been proposed to solve the problem, including $\ell_1$ convex relaxation approach [37] and the well-known Matching Pursuit

Family (MPF) algorithms that work in an iteratively greedy way [38].

For data compression, extensive efforts have been made to improve the typical sparse model. On one hand, it is known that the dictionary is critical to the coding performance, which is required to be adaptive to the image content. On the other hand, it would be rather expensive to represent an adaptive dictionary. Therefore, the idea of the double sparsity framework was presented [39] and developed [40], where both the dictionary and the coefficients were required to be sparse and easy to code. Based on that, the adaptively learnt dictionary was encoded in order to achieve efficiency and adaptivity simultaneously [41]. Furthermore, an online dictionary learning algorithm for intra-frame video coding was proposed [42], where the dictionary was dynamically updated across video frames and only the dictionary changes were encoded in the stream. However, the coding efficiency of this kind of approaches was highly limited due to the considerable bitrate consumptions of the dictionary.

Therefore, for most of the existing works, a global dictionary was offline trained and shared in the encoder and decoder sides. Typical works attempted to train a content-aware dictionary for the facial image compression [43]–[45], screen image compression [46] and fingerprint image compression [47]. For general image compression, several approaches have been proposed to improve the representation ability of the dictionary. Boosted dictionary training algorithm was developed for this purpose [48]. In [49], the RLS dictionary learning algorithm [35] was employed in the 9/7 wavelet domain. Recently, a new sparse dictionary learning model was proposed by imposing a compressibility constraint on the coefficients [25]. In [50], an image is partitioned to basis blocks and non-basis blocks, and non-basis blocks are compressed using the dictionary trained by basis blocks. However, there is still much room to improve the compression efficiency of the sparse model. Particularly, in conventional sparse models, the sparsity constraint term can not well reflect the actual coding bits of the coefficients. This motivates us to propose the GVCSR model for the purpose of rate-distortion optimized sparse representation.

## III. GLOBALLY VARIANCE-CONSTRAINED SPARSE REPRESENTATION

In this section, we firstly present the GVCSR model. The effective methods for sparse coding and dictionary learning are subsequently introduced. Finally, some implementation issues and their solutions are introduced.

### A. Rate-Distortion Optimized Sparse Representation

For data compression, the objective function that takes the coding rate into consideration is formulated as follows,

$$\arg\min_{A,D} \left\{ \frac{1}{2} \|S - DA\|_F^2 + \lambda \cdot r(A) \right\},$$
$$\text{s.t. } \|A_i\|_0 \leqslant L, \forall i \in \{1, \cdots, K\},$$
$$\|D_j\|_2^2 \leqslant 1, \forall j \in \{1, 2, \ldots, M\}, \quad (3)$$

where $S \in \mathbb{R}^{N \times K}$ is the stack of input vectors $S_i$. $A \in \mathbb{R}^{M \times K}$ is the stack of the corresponding sparse coefficient vectors $A_i$, and $K$ denotes the number of the input samples. $r(\cdot)$ is a function that can represent the coding rate of the coefficients. This formula can be interpreted as the rate-distortion optimization in common image/video compression [18], where $\lambda$ controls the relative importance between the rate and distortion. The $\ell_0$ norm of the coefficients is critical in order to obtain a sparse approximation. Subsequently, the problem turns to how to accurately estimate the coding rate and efficiently optimize it in sparse coding and dictionary learning.

Based on the *Shannon*'s information theory, the entropy of a data source indicates the average number of bits required to represent it. However, it is difficult to estimate the probability density function of coefficients and formulate the entropy minimization problem. Fortunately, the entropy can be bounded by the data variance according to the *Shannon* entropy inequality [51],

$$H(A) \leqslant \log\left(\sqrt{2\pi e V(A)}\right), \quad (4)$$

where $H(A)$ and $V(A)$ indicate the entropy and the variance of coefficients, respectively. Note that the inequality is tight as the equality holds for Gaussian distribution. Actually, the distribution of sparse coefficients is always assumed to be Laplacian distribution in the literatures (e.g. in [52]), which is close to a Gaussian distribution. The entropy of a Laplacian distribution can be calculated by,

$$H_L(A) = \log\left(\sqrt{2\pi e V(A)}\right) - \log\left(\sqrt{\pi/e}\right), \quad (5)$$

where $\log\left(\sqrt{\pi/e}\right)$ is a pretty small positive, indicating the entropy of Laplacian distribution can be tightly bounded by Eqn. (4).

To further validate this, we generate Laplacian distributed data and encode them by Huffman coding. The relationship between the data variance and the actual coding bits is plotted in Fig. 1. From the figure, one can observe that the variance exhibits a strong relationship with the coding bitrate. It convinces us to use the variance as the rate estimation term.

Another benefit of using variance is that the variance can be estimated by,

$$V(A) = tr\left(AZA^T\right), \quad (6)$$

where

$$Z = \begin{pmatrix} K-1 & \cdots & -1 \\ \vdots & \ddots & \vdots \\ -1 & \cdots & K-1 \end{pmatrix} \in \mathbb{R}^{K \times K}. \quad (7)$$

The diagonal elements in $Z$ equal to $K - 1$ and others equal to $-1$.

Therefore, we propose to minimize the variance as an estimation of entropy. This is reasonable since minimizing variance encourages the coefficients to be closer to each other, which is more friendly to compression. The objective function in (3) can be formulated as follows,

$$\arg\min_{A,D} \left\{ \frac{1}{2} \|S - DA\|_F^2 + \frac{\beta}{2} tr\left(AZA^T\right) \right\},$$
$$\text{s.t. } \|A_i\|_0 \leqslant L, \forall i \in \{1, \cdots, K\},$$
$$\|D_j\|_2^2 \leqslant 1, \forall j \in \{1, 2, \ldots, M\}, \quad (8)$$

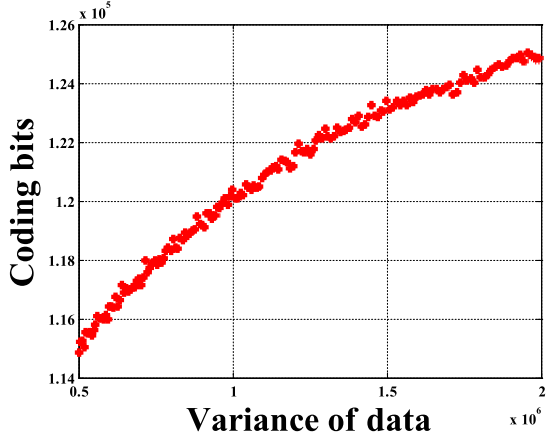

Fig. 1. Relationship between the data variance and the actual coding bits.

where $\beta/2$ is introduced for computational convenience. Generally, the $\ell_0$ norm constraint can be approximately solved by the Lagrangian method,

$$\arg\min_{A,D} \left\{ \frac{1}{2}\|S - DA\|_F^2 + \alpha\|A\|_0 + \frac{\beta}{2} tr\left(AZA^T\right) \right\},$$
$$\text{s.t. } \|D_i\|_2^2 \leqslant 1, \forall i \in \{1, 2, \ldots, M\}. \quad (9)$$

To solve the non-convex optimization problem effectively, a practical relaxation is to split the problem into two separable parts and update $A$ and $D$ alternately, i.e. the GVCSR based sparse coding and GVCSR based dictionary learning.

### B. GVCSR Based Sparse Coding

The GVCSR based sparse coding given the dictionary $D$ is a subproblem of (9), which can be formulated as follows,

$$A = \arg\min_{A} \left\{ \frac{1}{2}\|S - DA\|_F^2 + \alpha\|A\|_0 + \frac{\beta}{2} tr\left(AZA^T\right) \right\}. \quad (10)$$

To solve this, the Alternating Direction Method of Multipliers (ADMM) [29], [53] is employed in this work. The ADMM was originally introduced in the early 1970s [54], and has been widely used in machine learning, computer vision and signal processing [29], [55]–[57]. Its convergence properties for both convex and non-convex problems were also intensively addressed [29], [58]. It is known that the ADMM is efficient in dealing with the following problem,

$$\arg\min_{x,y} \{f(x) + g(y)\}, \text{ s.t. } \mathcal{B}(x) + \mathcal{C}(y) - d = 0, \quad (11)$$

where $x$, $y$ and $d$ could be either vectors or matrices. $\mathcal{B}$ and $\mathcal{C}$ are linear operators that define the constraint function. By introducing Lagrangian multiplier vector $R$, the augmented Lagrangian function can be formed as follows,

$$\zeta_\mu(x, y, R) = f(x) + g(y) + \langle \mathcal{B}(x) + \mathcal{C}(y) - d, R\rangle$$
$$+ \frac{\mu}{2}\|\mathcal{B}(x) + \mathcal{C}(y) - d\|_2^2, \quad (12)$$

where $\mu > 0$ is the penalty parameter. The ADMM updates the estimation of $x, y, R$ via solving the following problems alternately,

$$\begin{cases} x^{k+1} = \arg\min_x \zeta_\mu(x, y^k, R^k) \\ y^{k+1} = \arg\min_y \zeta_\mu(x^{k+1}, y, R^k) \\ R^{k+1} = R^k + \mu\left(\mathcal{B}(x^{k+1}) + \mathcal{C}(y^{k+1}) - d\right). \end{cases} \quad (13)$$

To solve (10), we first introduce two auxiliary variables $J$ and $G$ in order to fit the form in (11),

$$\arg\min_{A,J,G} \left\{ \frac{1}{2}\|S - DJ\|_F^2 + \alpha\|A\|_0 + \frac{\beta}{2} tr\left(GZG^T\right) \right\},$$
$$\text{s.t. } A = J, \quad A = G, \quad (14)$$

by setting the parameters in (11) as follows,

$$\begin{cases} x = A \\ y = \begin{bmatrix} J \\ G \end{bmatrix} \\ f(x) = \alpha\|A\|_0 \\ g(y) = \frac{1}{2}\|S - DJ\|_F^2 + \frac{\beta}{2} tr\left(GZG^T\right) \\ \mathcal{B}(x) = \begin{bmatrix} A \\ A \end{bmatrix}, \quad \mathcal{C}(y) = \begin{bmatrix} -J \\ -G \end{bmatrix}, \quad d = 0, \end{cases} \quad (15)$$

Then, the augmented Lagrangian function of (14) can be formulated by,

$$\zeta_\mu(A, J, G, R) = \frac{1}{2}\|S - DJ\|_F^2 + \alpha\|A\|_0 + \frac{\beta}{2} tr\left(GZG^T\right)$$
$$+ \langle A - J, R_0\rangle + \frac{\mu}{2}\|A - J\|_F^2$$
$$+ \langle A - G, R_1\rangle + \frac{\mu}{2}\|A - G\|_F^2, \quad (16)$$

where $R \triangleq [R_0; R_1]$ is the Lagrange multiplier matrix.

The variables $A$, $J$ and $G$ can be alternately updated by minimizing the augmented Lagrangian function $\zeta$ with other variables fixed. In this model, each variable can be updated with a closed form solution. Regarding $A$, it can be updated as follows,

$$A = \arg\min_A \left\{ \alpha\|A\|_0 + \langle A - J, R_0\rangle + \frac{\mu}{2}\|A - J\|_F^2 \right.$$
$$\left. + \langle A - G, R_1\rangle + \frac{\mu}{2}\|A - G\|_F^2 \right\}$$
$$= \arg\min_A \left\{ \alpha\|A\|_0 + \frac{\mu}{2}\left\|A - J + \frac{R_0}{\mu}\right\|_F^2 \right.$$
$$\left. + \frac{\mu}{2}\left\|A - G + \frac{R_1}{\mu}\right\|_F^2 \right\}$$
$$= H_{\sqrt{\alpha/\mu}}\left\{ \frac{1}{2}\left(J + G - \frac{R_0 + R_1}{\mu}\right) \right\}, \quad (17)$$

where

$$H_\varepsilon(X) \triangleq \begin{pmatrix} h_\varepsilon(X_{11}) & \cdots & h_\varepsilon(X_{1n}) \\ \vdots & \ddots & \vdots \\ h_\varepsilon(X_{m1}) & \cdots & h_\varepsilon(X_{mn}) \end{pmatrix}, \quad (18)$$

and

$$h_\varepsilon(x) \triangleq \begin{cases} x, & \text{if } |x| > \varepsilon \\ 0, & \text{if } |x| \leqslant \varepsilon \end{cases} \quad (19)$$

is a hard threshold operator. With respect to $J$ and $G$, we can update them as follows,

$$J = \arg\min_J \left\{ \frac{1}{2}\|S - DJ\|_F^2 + \langle A - J, R_0 \rangle + \frac{\mu}{2}\|A - J\|_F^2 \right\}$$

$$= \arg\min_J \left\{ \frac{1}{2}\|S - DJ\|_F^2 + \frac{\mu}{2}\left\|A - J + \frac{R_0}{\mu}\right\|_F^2 \right\}$$

$$= V_D \left( \Sigma_D^T \Sigma_D + \mu I \right)^{-1} V_D^T \left( D^T S + \mu A + R_0 \right), \quad (20)$$

$$G = \arg\min_G \left\{ \frac{\beta}{2} tr\left(GZG^T\right) + \langle A - G, R_1 \rangle + \frac{\mu}{2}\|A - G\|_F^2 \right\}$$

$$= \arg\min_G \left\{ \frac{\beta}{2} tr\left(GZG^T\right) + \frac{\mu}{2}\left\|A - G + \frac{R_1}{\mu}\right\|_F^2 \right\}$$

$$= (\mu A + R_1) V_Z (\beta \Sigma_Z + \mu I)^{-1} V_Z^T, \quad (21)$$

where $U_D \Sigma_D V_D^T$ and $U_Z \Sigma_Z V_Z^T$ are the full Singular Value Decomposition (SVD) of $D$ and $Z$, respectively. Finally, the Lagrangian multiplier $R_0$ and $R_1$ are updated,

$$R_0^{j+1} = R_0^j + \mu^j \left( A^{j+1} - J^{j+1} \right), \quad (22)$$

$$R_1^{j+1} = R_1^j + \mu^j \left( A^{j+1} - G^{j+1} \right), \quad (23)$$

where $j$ indicates the iteration times.

In previous ADMM approach [29], the penalty parameter $\mu$ is fixed. To accelerate the convergence, an updating strategy for the penalty parameter is proposed in [59], which can be formulated as follows,

$$\mu^{j+1} = \min\left(\rho \mu^j, \mu_{max}\right), \quad (24)$$

where $\mu_{max}$ is an upper bound of the penalty term. $\rho \geqslant 1$ is a constant.

The optimization process of the GVCSR based sparse coding performs iteratively and stops until convergence. Afterwards, the globally variance-constrained sparse coding can be achieved. The detailed procedure is presented in Algorithm 1.

### C. GVCSR Based Dictionary Learning

In order to make the dictionary learning consistent with the proposed sparse coding strategy, we further solve the dictionary updating problem based on the GVCSR model of Eqn. (9). To solve this, the proposed algorithm updates the sparse coefficients and the dictionary alternately and iteratively. Specifically, for each iteration, it first updates sparse coefficients by applying GVCSR based sparse coding algorithm, as discussed in Section III-B. After the convergence of $A$, the dictionary $D$ can be updated with the optimized $A$ as follows,

$$D = \arg\min_D \left\{ \frac{1}{2}\|S - DA\|_F^2 \right\}$$
$$\text{s.t. } \|D_i\|_2^2 \leqslant 1, \quad \forall i \in \{1, 2, \ldots, M\}. \quad (25)$$

This can be solved via performing SVD on the residuals [33]. Then, the parameters $A$, $J$, $G$ should be further updated

---

**Algorithm 1** GVCSR Based Sparse Coding in (10)

**Input:**
- Input data $S$
- Dictionary $D$
- Parameters $\alpha > 0$ and $\beta > 0$

**Output:**
- Sparse representation coefficients $A$

1 **Initialization**:
2 Set $A^0 = J^0 = G^0 = D^\dagger S, \mu^0 = 1e-2, \mu_{max} = 1e8, \rho = 1.2, R_0 = R_1 = 0, \epsilon = 1e-5$, and $j = 0$;
3 **while** *not convergence* **do**
4     Fix $J^j$ and $G^j$ to update $A^{j+1}$ by (17);
5     Fix $A^{j+1}$ and $G^j$ to update $J^{j+1}$ by (20);
6     Fix $A^{j+1}$ and $J^{j+1}$ to update $G^{j+1}$ by (21);
7     Update Lagrange multipliers:
       $R_0^{j+1} = R_0^j + \mu^j \left( A^{j+1} - J^{j+1} \right)$,
       $R_1^{j+1} = R_1^j + \mu^j \left( A^{j+1} - G^{j+1} \right)$;
8     Update penalty parameter $\mu$:
       $\mu^{j+1} = \min\left(\rho \mu^j, \mu_{max}\right)$;
9     $j \leftarrow j + 1$;
10    Check convergence: if $\|A^j - J^j\|/\|A^j\| \leqslant \epsilon$ and $\|A^j - G^j\|/\|A^j\| \leqslant \epsilon$ and $\|A^j - A^{j-1}\|/\|A^j\| \leqslant \epsilon$, then stop;
11 **end**

---

according to the new dictionary, while the Lagrangian multipliers $R_0$ and $R_1$ are reset to zero vectors. Moreover, the penalty parameter $\mu$ is updated as follows,

$$\mu^{j+1} = \frac{\kappa \cdot \alpha}{\min |A_{nz}^j|}, \quad (26)$$

where $A_{nz}^j$ denotes all the nonzero elements in $A^j$. $\kappa \geqslant 1$ is a parameter that guides how largely the $\|A\|_0$ would change in the next iteration when updating $A$. $\kappa$ is empirically set as 4 in this work. The procedure of the GVCSR based dictionary learning scheme is detailed in Algorithm 2.

### D. Implementation Issues

In (20) & (21), the full-SVD should be performed when updating the variables. The computational complexity is $O(N^3)$. It is unpractical since the dimension of matrix $Z \in \mathbb{R}^{K \times K}$ can be pretty high. Therefore, we propose a fast full-SVD algorithm for the matrix $Z$.

Let $V_Z = \{v_z^1, \cdots, v_z^K\}$ be the singular vectors of $Z$, and the corresponding singular values are $\{\sigma_z^1, \cdots, \sigma_z^K\}$. It is easy to derive that $Z$ has only two singular values, i.e. 0 and $K$. The singular value 0 corresponds to an all-one singular vector, and the singular value $K$ corresponds to $K - 1$ singular vectors. Thus we have

$$\Sigma_Z = \begin{pmatrix} 0 & & & \\ & K & & \\ & & \ddots & \\ & & & K \end{pmatrix} \in \mathbb{R}^{K \times K}. \quad (27)$$

**Algorithm 2** GVCSR Based Dictionary Learning in (9)

**Input:**
- Input data $S$
- Parameters $\alpha > 0$ and $\beta > 0$

**Output:**
- Trained dictionary $D$
- Sparse representation coefficients $A$

1 **Initialization**:
2 Initialize $D$;
3 Set $A^0 = J^0 = G^0 = D^\dagger S, \mu^0 = 1e-2, \mu_{max} = 1e8, \rho = 1.2, R_0^0 = R_1^0 = 0, \epsilon = 1e-5,$ and $j=0$;
4 **while** *not convergence* **do**
5    **Updating Sparse Coefficients**:
6    Fix $J^j$ and $G^j$ to update $A^{j+1}$ by (17);
7    Fix $A^{j+1}$ and $G^j$ to update $J^{j+1}$ by (20);
8    Fix $A^{j+1}$ and $J^{j+1}$ to update $G^{j+1}$ by (21);
9    Update Lagrange multipliers:
$$R_0^{j+1} = R_0^j + \mu^j \left( A^{j+1} - J^{j+1} \right),$$
$$R_1^{j+1} = R_1^j + \mu^j \left( A^{j+1} - G^{j+1} \right);$$
10    Update penalty parameter $\mu$:
$$\mu^{j+1} = \min\left( \rho\mu^j, \mu_{max} \right);$$
11    $j \leftarrow j+1$;
12    Check convergence:
13    **if** $\|A^j - J^j\|/\|A^j\| \leqslant \epsilon$ and $\|A^j - G^j\|/\|A^j\| \leqslant \epsilon$ and $\|A^j - A^{j-1}\|/\|A^j\| \leqslant \epsilon$ **then**
14      **Updating Dictionary**:
15      Updating $D$ by solving (25);
16      $A^{j+1} = J^{j+1} = G^{j+1} = D^\dagger S$;
17      $R_0^{j+1} = R_1^{j+1} = 0$;
18      Update $\mu^{j+1}$ by (26);
19      $j \leftarrow j+1$;
20    **end**
21 **end**

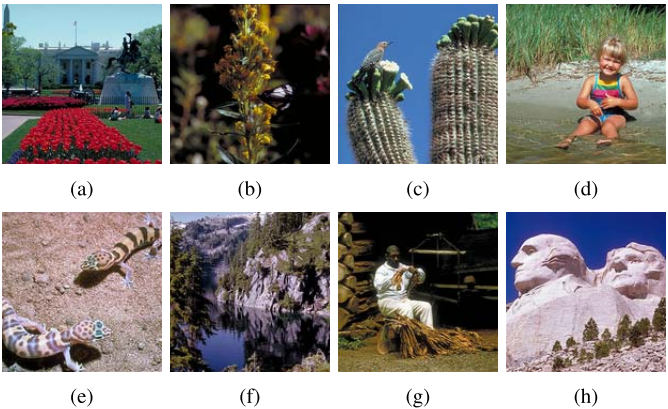

Fig. 2. Illustration of the training images from CSIQ database [60]. (a) 1600. (b) Flower. (c) Cactus. (d) Child. (e) Geckos. (f) Lake. (g) Roping. (h) Rushmore.

Based on the principle of the SVD, the corresponding singular vectors must satisfy,

$$\langle v_z^i, v_z^j \rangle = 0 \text{ and } \langle v_z^i, (1, \cdots, 1)^T \rangle = 0,$$
$$\forall i \neq j \in \{2, \cdots, K\}, \quad (28)$$

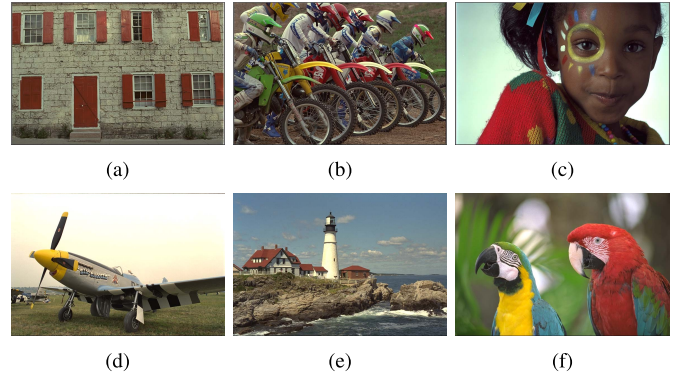

Fig. 3. Illustration of the testing images from Kodak database [61]. (a) Building. (b) Bikes. (c) Girl. (d) Plane. (e) Lighthouse. (f) Macaws.

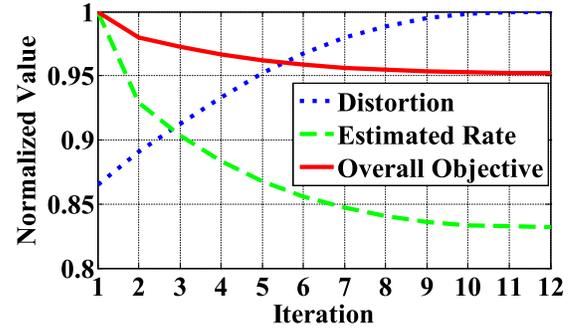

Fig. 4. Changing curves of distortion (measured by mean squared error), estimated rate and the overall objective value in terms of iterations, which are plotted by the blue dotted line, green dashed line and red solid line, respectively. The horizontal axis denotes the iteration numbers. Note that the three curves are normalized to the same range to show them in one figure.

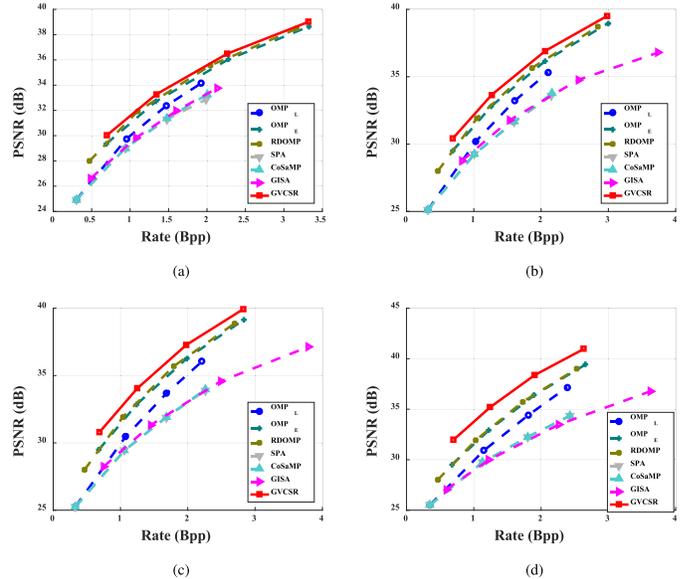

Fig. 5. Rate-distortion performance comparisons with other sparse coding algorithms, in terms of dictionaries with different values of $\gamma$, where $\gamma$ is defined as $\gamma \triangleq M/N$. (a) Completeness $\gamma = 4$. (b) Completeness $\gamma = 8$. (c) Completeness $\gamma = 12$. (d) Completeness $\gamma = 16$.

and $v_z^1 = (1, \cdots, 1)^T$ is the all-one vector. Equivalently, the problem can be reformulated to find an orthogonal matrix $V_Z$ given $v_z^1$. Consequently, we can construct the singular vector matrix as follows,

$$V_z = \begin{pmatrix} 1 & 1 & 1 & 1 & 1 & 1 \\ 1 & -1 & 1 & \vdots & 1 & 1 \\ \vdots & 0 & -2 & 1 & \vdots & 1 \\ 1 & \vdots & 0 & \ddots & 1 & \vdots \\ 1 & 0 & \vdots & 0 & 2-K & 1 \\ 1 & 0 & 0 & 0 & 0 & 1-K \end{pmatrix} \in \mathbb{R}^{K \times K}, \quad (29)$$

followed by a normalization process to satisfy $\|v_z^i\|_2^2 = 1$ for $i \in \{1, \cdots, K\}$. Compared with the normal full-SVD decomposition, this method is much faster as its complexity can be reduced to $O(N)$.

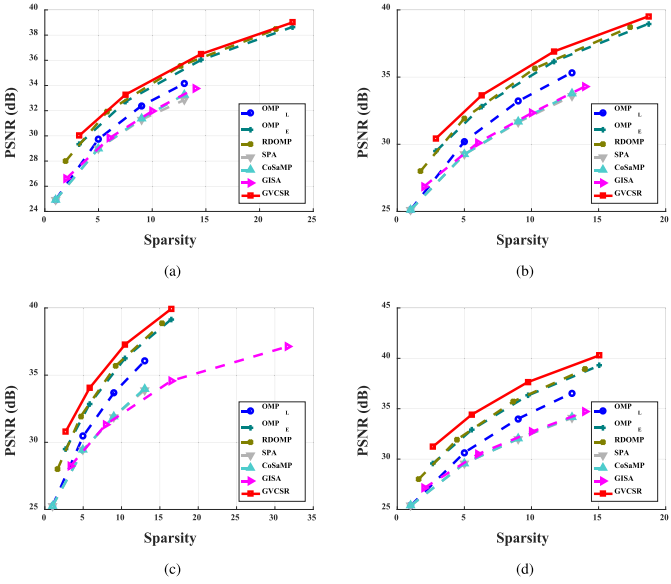

Fig. 6. Sparsity-distortion performance comparisons with other sparse coding algorithms, in terms of dictionaries with different values of $\gamma$, where $\gamma$ is defined as $\gamma \triangleq M/N$. The horizontal axis indicates the average number of nonzero coefficients. (a) Completeness $\gamma = 4$. (b) Completeness $\gamma = 8$. (c) Completeness $\gamma = 12$. (d) Completeness $\gamma = 16$.

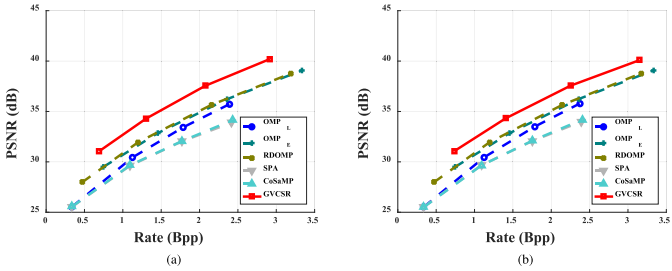

Fig. 7. Rate-distortion performance comparisons with other sparse coding algorithms. Note that the sparse coding stage in dictionary learning is replaced by the same sparse coding scheme as calculating the coefficients for entropy coding. The basic dictionary learning methods are (a) K-SVD and (b) MOD, respectively.

According to the convergence theory of the ADMM (please refer to the Assumption 1 and Theorem 1 on Pages 5-10 of [58]), the following conclusion can be achieved.

*Theorem 1:* If the penalty parameter $\mu$ is chosen to be larger than $\{\sqrt{2}\lambda_{max}(D^T D)\}$ (where $\lambda_{max}(\bullet)$ denotes the largest eigenvalue) and the sequence $\{(J^k, A^k, G^k)\}$ generated from the ADMM has a cluster point $\{(J^*, A^*, G^*)\}$, then $A^*$ is a critical point of (14).

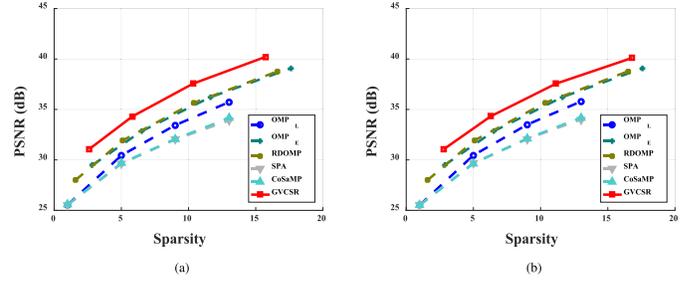

Fig. 8. Sparsity-distortion performance comparisons with other sparse coding algorithms. Note that the sparse coding stage in dictionary learning is replaced by the same sparse coding scheme as calculating the coefficients for entropy coding. The basic dictionary learning methods are (a) K-SVD and (b) MOD, respectively.

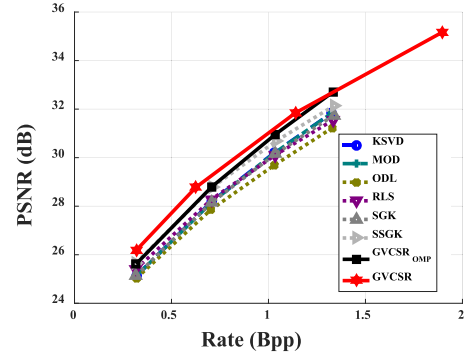

Fig. 9. Rate-distortion performance comparisons with the state-of-the-art dictionary learning algorithms, including MOD [31], K-SVD [33], ODL [34], RLS [35], SGK [36] and Sparse-SGK (SSGK) [40]. During sparse coding process, all other algorithms including $GVCSR_{OMP}$ employ the OMP method while $GVCSR$ utilizes the GVCSR based sparse coding method.

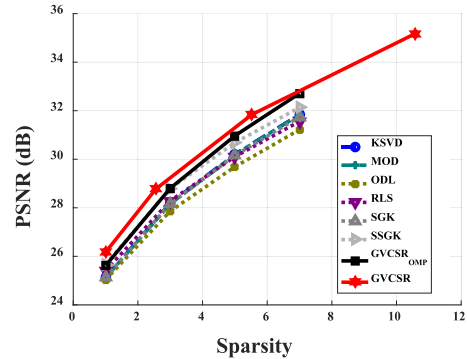

Fig. 10. Sparsity-distortion performance comparisons with the state-of-the-art dictionary learning algorithms, including MOD [31], K-SVD [33], ODL [34], RLS [35], SGK [36] and Sparse-SGK (SSGK) [40]. During sparse coding process, all other algorithms including $GVCSR_{OMP}$ employ the OMP method while $GVCSR$ utilizes the GVCSR based sparse coding method.

## IV. EXPERIMENTAL RESULTS

In this section, the performance of the proposed GVCSR model is validated from two perspectives. First, we compare the rate-distortion performance of the proposed method with other sparse coding and dictionary learning algorithms in image representation. Second, we apply it to the practical image set compression.

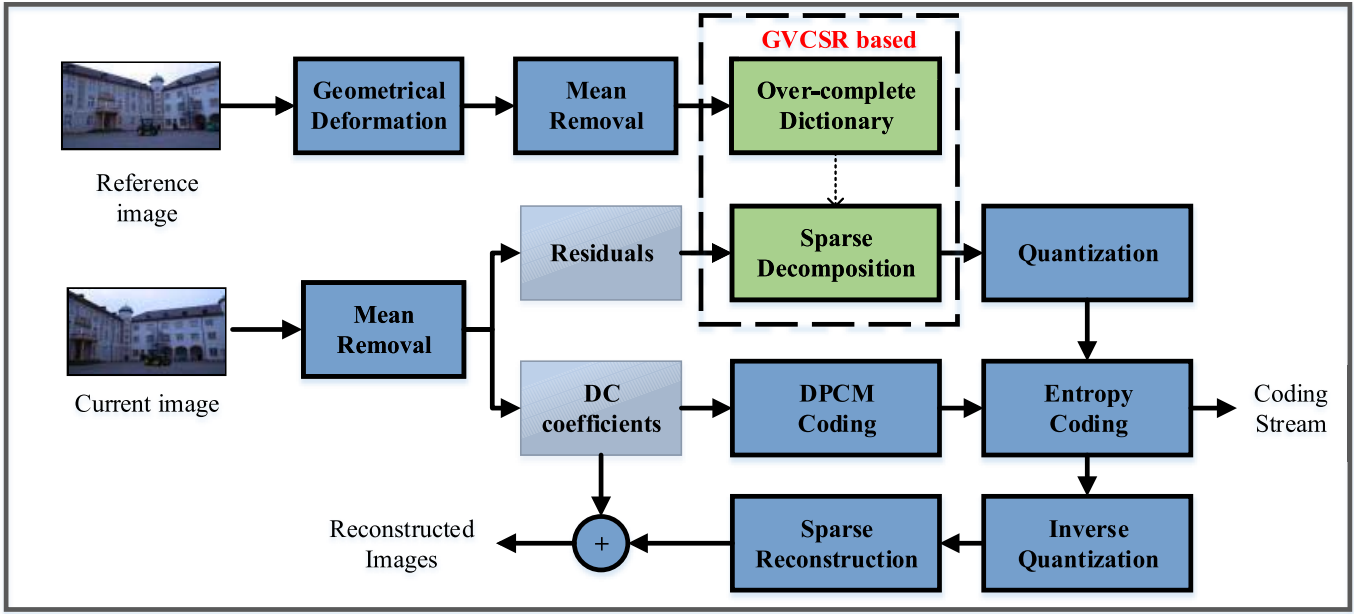

Fig. 11. Image set compression framework based on the GVCSR model. The dictionary is trained using the reference image by applying the GVCSR based dictionary learning. The GVCSR based sparse coding in terms of the learnt dictionary is applied for coding the current image.

### A. GVCSR for Image Representation

In the experiments, the CSIQ dataset [60] (containing 30 original images) is utilized for training dictionary, and Kodak database [61] (containing 24 original images) is employed for testing. Some examples are shown in Figs. 2 & 3. Each image is partitioned into $8 \times 8$ non-overlapped blocks. Note that in the experiments all the images are converted to gray scale.

First, the changing tendencies of the distortion, rate and the overall objective function during the ADMM iterations are shown in Fig. 4. The rate is estimated by the variance of sparse coefficients as described in Section III-A, and the overall objective value is calculated by (9). To show three curves in one figure, the values are normalized to a same range. From the figure, one can observe that the proposed method can achieve better tradeoff in terms of the overall objective by greatly decreasing the coding rate with slightly increased distortion. As a result, the coding performance can be improved.

Then, we compare the GVCSR based sparse coding algorithm with other sparse coding approaches. The first method for comparison is the standard OMP algorithm [62], where the iteration process stops until the $\ell_0$ norm of coefficients reaches the limited value $L$. The second one is similar but the stop criterion is determined by the error energy,

$$A_i = \arg\min_{A_i} \|A_i\|_0, \quad \text{s.t.} \ \|S_i - DA_i\|_2^2 < \epsilon. \quad (30)$$

These two methods are denoted as $OMP_L$ and $OMP_E$, respectively. The RDOMP method [20]–[22], which also considers rate constraint by the probability distribution of coefficients, is the third competitive method. Three other state-of-the-art methods of sparse coding, CoSaMP [63], SPA [64], and GISA [65], are also used for comparison.

For fair comparison, the K-SVD algorithm [33] is employed for dictionary training. In Fig. 5, the rate-distortion comparisons are illustrated in terms of different values of completeness, where the completeness is defined as $\gamma \triangleq \frac{M}{N}$ considering the dictionary $D \in \mathbb{R}^{N \times M}$. Since the coefficients are sparse, the run-length encoding method is employed to code the values and positions of the nonzero sparse coefficients. It can be observed that the rate-distortion performance is significantly improved by the proposed algorithm. This may benefit from the global optimization of the proposed method that jointly considers the distortion and coding rate, while other methods are based on local optimization. The last three methods for comparison [63]–[65] show worse performance. This may be caused by the reason that they did not take rate term into consideration, as they were designed for signal recovery applications rather than compression. Another observation is that the improvements are more obvious for larger $\gamma$, where the coefficients are sparser. In such case, the matching pursuit based methods may suffer from their potential instability due to the increasing independence among coefficients, while the proposed method can effectively solve this by global optimization to significantly reduce the coding bits. In addition to rate-distortion curves, the sparsity-distortion curves are shown in Fig. 6, where the horizontal axis denotes the average number of nonzero coefficients. From these figures, we can have similar conclusions that the proposed method can obtain higher PSNR with the same sparsity level.

To further fairly compare the performance of different sparse coding algorithms, we conduct the following comparison experiments. First, we keep the original dictionary updating process as untouched (e.g. K-SVD or MOD). Second, the original sparse coding stage (typically the $OMP_L$) in dictionary learning is replaced by other sparse coding algorithms (e.g. $OMP_E$, RDOMP, SPA, CoSaMP and the proposed

TABLE II

Running Time Comparison Between GVCSR and Other Dictionary Learning Algorithms. All the Experiments Are Tested by Matlab R2016a on Windows PC With an Intel Core CPU i7-4770 @3.40 GHz. The Time Is Measured by Second Per Iteration

|  | K-SVD | MOD | ODL | RLS | SGK | SSGK | GVCSR |
|---|---|---|---|---|---|---|---|
| **Time** | 2.23 | 0.59 | 1.88 | 5.16 | 0.02 | 0.01 | 2.72 |

GVCSR). We then use the same sparse coding algorithm when calculating the coefficients before entropy coding. The performance of using different sparse coding schemes can be compared in a fair way, since the same sparse coding algorithm is applied for dictionary updating and generating sparse coefficients. The rate-distortion and sparsity-distortion comparison results of using K-SVD and MOD methods are shown in Figs. 7 & 8, respectively. From these results, one can observe it more clearly that the proposed GVCSR method outperforms all the other sparse coding algorithms.

Finally, we compare the performance of the GVCSR based dictionary learning scheme with the state-of-the-art dictionary learning algorithms, including MOD [31], K-SVD [33], ODL [34], RLS [35], SGK [36] and Sparse-SGK (SSGK) [40]. To be fair, the initial dictionary in each algorithm is the same, which is randomly selected from the training set. The OMP algorithm is employed for sparse coding. Note that two versions of GVCSR based method are compared. The first one denoted as $GVCSR_{OMP}$ indicates that the OMP algorithm is used for sparse coding instead of the GVCSR based algorithm. Comparing $GVCSR_{OMP}$ with other competitive methods is fair because they use the same sparse coding algorithm. However, in this case, the GVCSR based dictionary becomes suboptimal since the targets of dictionary learning and sparse coding are inconsistent. Therefore, we evaluate the second method denoted as $GVCSR$, where GVCSR based sparse coding is employed. The average results of rate-distortion and sparsity-distortion performance are shown in Figs. 9 & 10, respectively. One can see that the $GVCSR_{OMP}$ scheme can achieve significant improvements over other competitors even though it is suboptimal, and the $GVCSR$ scheme can bring more gains. It demonstrates that the proposed variance constraint is more capable of estimating coding bits than a sparsity constraint alone. We also compare the time complexity of the proposed method with other schemes in Table II. The runtime of the GVCSR is close to the K-SVD and much faster than the RLS.

### B. Image Set Compression Based on GVCSR Model

In this subsection, we apply the proposed algorithm on the application of image set compression. Generally, an image set contains the same object of interest captured from different luminance conditions and viewpoints, indicating strong correlations among images. However, traditional single image compression methods only explore intra-image dependencies while the inter-image dependencies have been ignored. To utilize such kind of correlations, several approaches have been proposed by explicitly performing inter-image predictions where similar images are considered as video frames [66]–[73]. Their basic idea was to explore inter-image correlations by geometric transformation based prediction between similar images.

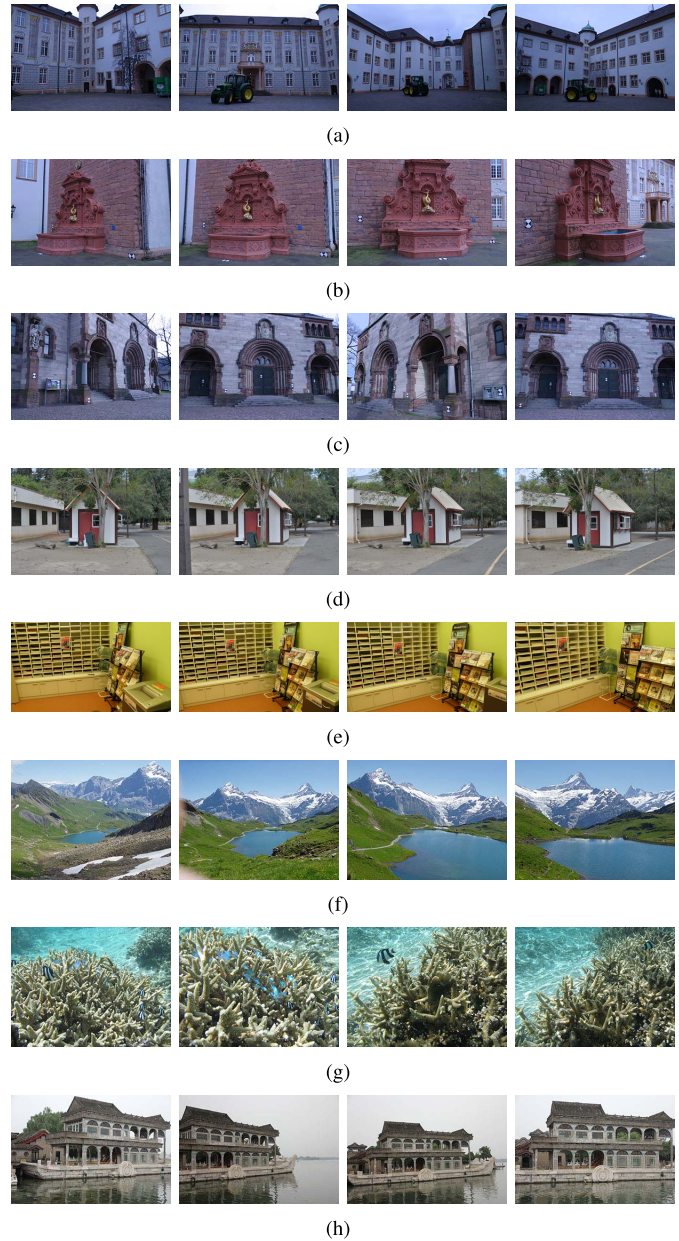

Fig. 12. Examples of the image sets [69] [74] and [75]. (a) CastleEntry. (b) Fountain. (c) Herzjesu. (d) UCD. (e) MallRoom. (f) Lakes. (g) CoralReef. (h) RockBoat.

The proposed scheme implicitly eliminates inter-image redundancies by training dictionary from similar images, instead of directly applying inter-image prediction. The coding framework is shown in Fig. 11, which consists of three stages, i.e. the dictionary training, sparse coding and coefficients compression. During the dictionary training stage, the current image will find a similar image as reference in the image set. The reference structure is determined by the Minimum Spanning Tree (MST) algorithm, where the similarity between two images is measured by the matched feature distance [71]. Since the MST is calculated based on original images, the MST structure is encoded with negligible bits to make sure the receiver can correctly decode the images. Then the reference image is used to train the over-complete dictionary by the proposed GVCSR algorithm. Since the reference image is available on both encoder and decoder sides, the dictionary can be trained identically on both sides while avoids transmitting overhead. Before training, the geometrical transformation is applied to the reference image for better exploring the inter-image correlations [69]. The training samples are $\sqrt{N} \times \sqrt{N}$ patches extracted from the transformed image. Regarding the sparse decomposition, the GVCSR based scheme is applied. The compression scheme can be further split into two parts. First, the Direct-Current (DC) component of each image patch is encoded by Differential Pulse Code Modulation (DPCM) coding, where the DC value of current block can be predicted from its left, top, top-left and top-right blocks. Second, the sparse coefficients for representing the residuals after mean removal are scalar-quantized and compressed by run-length coding method with entropy coding. The decoder side reconstructs the images using the decoded sparse coefficients and the DC values.

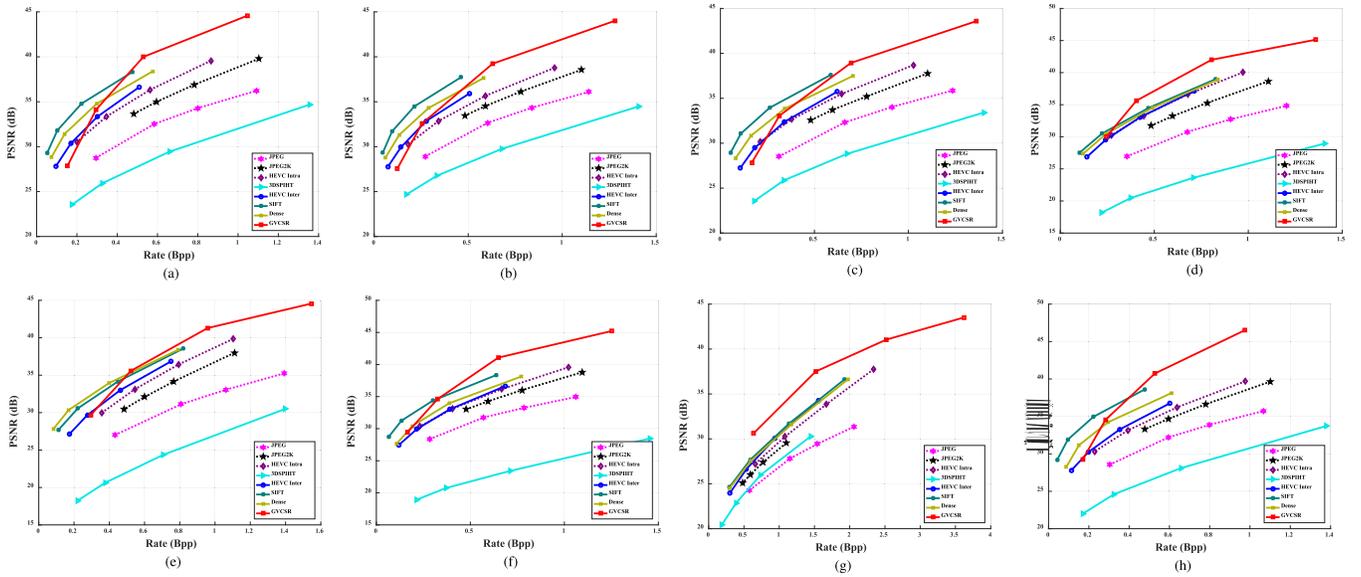

Fig. 13. Image set compression: RD performance comparisons with the state-of-the-art image codecs (JPEG, JPEG2000, HEVC intra [76]), video coding based methods (3DSPIHT [77], HEVC inter [76]) and image set coding schemes (SIFT [71], Dense [73]). (a) CastleEntry. (b) Fountain. (c) Herzjesu. (d) UCD. (e) MallRoom. (f) Lakes. (g) CoralReef. (h) RockBoat.

In this experiments, the dictionary is trained from $8 \times 8$ image patches, and its completeness value is set to be $\gamma = 14$, indicating the dictionary has $8 \times 8 \times 14 = 896$ bases. Test image sets are downloaded from public databases [69] [74] and [75], and part of them are shown in Fig. 12.

Fig. 13 demonstrates the Rate-Distortion (RD) curves of the proposed method comparing with the popular image compression standards (JPEG, JPEG2000, HEVC intra coding [76]), video compression based methods (3D-SPIHT [77], HEVC inter coding [76]) and two state-of-the-art image set compression algorithms ("SIFT" [71] and "Dense" [73]). It is worth noting that for fair comparisons, the HEVC configurations are modified by turning off the deblocking and sample adaptive offset filters and restricting the block partition depth to be 2 (i.e., $8 \times 8$ and $16 \times 16$ blocks). The proposed method utilizes the GVCSR based sparse coding and GVCSR based dictionary learning consistently. It should also be noted that the RD curves are averaged across all similar images in a set. From the results, we can observe that the proposed GVCSR model can obtain superior RD performance over traditional image/video codecs, since they do not utilize or weakly utilize inter-image correlations among similar images. Compared with the two image set coding schemes, the proposed method can provide competitive and even better performance at higher bitrate. The results encourage us to use the GVCSR method for practical compression applications.

## V. CONCLUSION

In this work, we present a novel Globally Variance-Constrained Sparse Representation (GVCSR) for rate-distortion joint optimization. To achieve this goal, a variance-constraint term that can accurately predict the coding rate of the sparse coefficients is introduced into the optimization process. Subsequently, we propose to use the Alternating Direction Method of Multipliers (ADMM) to effectively solve this model. In this manner, the rate-distortion jointly optimized sparse representation can be achieved, leading to higher compression efficiency. Furthermore, experimental results have shown that the GVCSR model can achieve better RD performance comparing with the state-of-the-art sparse coding and dictionary learning algorithms. We further demonstrate the effectiveness of the proposed method in image set compression, and better performance over competitive algorithms has been achieved.

ZHANG *et al.*: GVCSR AND ITS APPLICATION IN IMAGE SET CODING 3763[2] D. Taubman and M. Marcellin, *JPEG2000 Image Compression Fundamentals, Standards and Practice: Image Compression Fundamentals, Standards and Practice*, vol. 642. New York, NY USA: Springer, 2012.

[3] B. A. Olshausen and D. J. Field, "Emergence of simple-cell receptive field properties by learning a sparse code for natural images," *Nature*, vol. 381, no. 6583, pp. 607–609, 1996.

[4] M. Elad, *Sparse and Redundant Representations: From Theory to Applications in Signal and Image Processing*. New York, NY USA: Springer, 2010.

[5] J. Wright, Y. Ma, J. Mairal, G. Sapiro, T. S. Huang, and S. Yan, "Sparse representation for computer vision and pattern recognition," *Proc. IEEE*, vol. 98, no. 6, pp. 1031–1044, Jun. 2010.

[6] A. Hyvärinen, J. Hurri, and P. O. Hoyer, *Natural Image Statistics: A Probabilistic Approach to Early Computational Vision*, vol. 39. London, U.K.: Springer-Verlag, 2009.

[7] X. Zhang, S. Wang, S. Ma, S. Liu, and W. Gao, "Entropy of primitive: A top-down methodology for evaluating the perceptual visual information," in *Proc. Vis. Commun. Image Process. (VCIP)*, Nov. 2013, pp. 1–6.

[8] X. Zhang, S. Wang, S. Ma, R. Xiong, and W. Gao, "Towards accurate visual information estimation with entropy of primitive," in *Proc. IEEE Int. Symp. Circuits Syst. (ISCAS)*, May 2015, pp. 1046–1049.

[9] S. Ma, X. Zhang, S. Wang, J. Zhang, H. Sun, and W. Gao, "Entropy of primitive: From sparse representation to visual information evaluation," *IEEE Trans. Circuits Syst. Video Technol.*, vol. 27, no. 2, pp. 249–260, Feb. 2015.

[10] M. Elad and M. Aharon, "Image denoising via sparse and redundant representations over learned dictionaries," *IEEE Trans. Image Process.*, vol. 15, no. 12, pp. 3736–3745, Dec. 2006.

[11] J. Zhang, C. Zhao, R. Xiong, S. Ma, and D. Zhao, "Image super-resolution via dual-dictionary learning and sparse representation," in *Proc. IEEE Int. Symp. Circuits Syst. (ISCAS)*, May 2012, pp. 1688–1691.

[12] J. Zhang, D. Zhao, and W. Gao, "Group-based sparse representation for image restoration," *IEEE Trans. Image Process.*, vol. 23, no. 8, pp. 3336–3351, Aug. 2014.

[13] J. Zhang, C. Zhao, D. Zhao, and W. Gao, "Image compressive sensing recovery using adaptively learned sparsifying basis via L0 minimization," *Signal Process.*, vol. 103, pp. 114–126, Oct. 2014.

[14] J. Zhang, D. Zhao, R. Xiong, S. Ma, and W. Gao, "Image restoration using joint statistical modeling in a space-transform domain," *IEEE Trans. Circuits Syst. Video Technol.*, vol. 24, no. 6, pp. 915–928, Jun. 2014.

[15] H.-W. Chang, H. Yang, Y. Gan, and M.-H. Wang, "Sparse feature fidelity for perceptual image quality assessment," *IEEE Trans. Image Process.*, vol. 22, no. 10, pp. 4007–4018, Oct. 2013.

[16] X. Zhang, S. Wang, K. Gu, T. Jiang, S. Ma, and W. Gao, "Sparse structural similarity for objective image quality assessment," in *Proc. IEEE Int. Conf. Syst., Man, Cybern. (SMC)*, Oct. 2015, pp. 1561–1566.

[17] T. Guha, E. Nezhadarya, and R. K. Ward, "Learning sparse models for image quality assessment," in *Proc. IEEE Int. Conf. Acoust., Speech, Signal Process.*, May 2014, pp. 151–155.

[18] G. J. Sullivan and T. Wiegand, "Rate-distortion optimization for video compression," *IEEE Signal Process. Mag.*, vol. 15, no. 6, pp. 74–90, Nov. 1998.

[19] J. Zhang, S. Ma, R. Xiong, D. Zhao, and W. Gao, "Image primitive coding and visual quality assessment," in *Advances in Multimedia Information Processing—PCM*. Berlin, Germany: Springer, 2012, pp. 674–685.

[20] M. Gharavi-Aikhansari, "A model for entropy coding in matching pursuit," in *Proc. Int. Conf. Image Process. (ICIP)*, vol. 1. Oct. 1998, pp. 778–782.

[21] P. Vandergheynst and P. Frossard, "Adaptive entropy-constrained matching pursuit quantization," in *Proc. IEEE Int. Conf. Image Process. (ICIP)*, vol. 2. Oct. 2001, pp. 423–426.

[22] T. Ryen, G. M. Schustertt, and A. K. Katsaggelos, "A rate-distortion optimal coding alternative to matching pursuit," in *Proc. IEEE Int. Conf. Acoust., Speech, Signal Process. (ICASSP)*, vol. 3. May 2002, pp. III-2177–III-2180.

[23] T. Dumas, A. Roumy, and C. Guillemot, "Shallow sparse autoencoders versus sparse coding algorithms for image compression," in *Proc. IEEE Int. Conf. Multimedia Expo Workshops (ICMEW)*, Jul. 2016, pp. 1–6.

[24] A. Rahmoune, P. Vandergheynst, and P. Frossard, "Sparse approximation using M-term pursuit and application in image and video coding," *IEEE Trans. Image Process.*, vol. 21, no. 4, pp. 1950–1962, Apr. 2012.

[25] M. Xu, S. Li, J. Lu, and W. Zhu, "Compressibility constrained sparse representation with learnt dictionary for low bit-rate image compression," *IEEE Trans. Circuits Syst. Video Technol.*, vol. 24, no. 10, pp. 1743–1757, Oct. 2014.

[26] J. Mairal, F. Bach, J. Ponce, G. Sapiro, and A. Zisserman, "Non-local sparse models for image restoration," in *Proc. IEEE Int. Conf. Comput. Vis.*, Sep./Oct. 2009, pp. 2272–2279.

[27] W. Dong, L. Zhang, and G. Shi, "Centralized sparse representation for image restoration," in *Proc. IEEE Int. Conf. Comput. Vis. (ICCV)*, Nov. 2011, pp. 1259–1266.

[28] X. Liu, D. Zhai, D. Zhao, and W. Gao, "Image super-resolution via hierarchical and collaborative sparse representation," in *Proc. Data Compress. Conf. (DCC)*, Mar. 2013, pp. 93–102.

[29] S. Boyd, N. Parikh, E. Chu, B. Peleato, and J. Eckstein, "Distributed optimization and statistical learning via the alternating direction method of multipliers," *Found. Trends Mach. Learn.*, vol. 3, no. 1, pp. 1–122, Jan. 2011.

[30] X. Zhang, S. Ma, Z. Lin, J. Zhang, S. Wang, and W. Gao, "Globally variance-constrained sparse representation for rate-distortion optimized image representation," in *Proc. Data Compress. Conf. (DCC)*, Apr. 2017, pp. 380–389.

[31] K. Engan, S. O. Aase, and J. Hakon Husoy, "Method of optimal directions for frame design," in *Proc. IEEE Int. Conf. Acoust., Speech, Signal Process.*, vol. 5. Mar. 1999, pp. 2443–2446.

[32] K. Engan, K. Skretting, and J. H. Husøy, "Family of iterative LS-based dictionary learning algorithms, ILS-DLA, for sparse signal representation," *Digit. Signal Process.*, vol. 17, no. 1, pp. 32–49, 2007.

[33] M. Aharon, M. Elad, and A. Bruckstein, "K-SVD: An algorithm for designing overcomplete dictionaries for sparse representation," *IEEE Trans. Signal Process.*, vol. 54, no. 11, pp. 4311–4322, Nov. 2006.

[34] J. Mairal, F. Bach, J. Ponce, and G. Sapiro, "Online learning for matrix factorization and sparse coding," *J. Mach. Learn. Res.*, vol. 11, pp. 19–60, Mar. 2010.

[35] K. Skretting and K. Egan, "Recursive least squares dictionary learning algorithm," *IEEE Trans. Signal Process.*, vol. 58, no. 4, pp. 2121–2130, Apr. 2010.

[36] S. K. Sahoo and A. Makur, "Dictionary training for sparse representation as generalization of K-means clustering," *IEEE Signal Process. Lett.*, vol. 20, no. 6, pp. 587–590, Jun. 2013.

[37] R. Tibshirani, "Regression shrinkage and selection via the Lasso," *J. Roy. Statist. Soc. B (Methodol.)*, vol. 58, no. 1, pp. 267–288, 1996.

[38] J. Z. Salvatierra, "New sparse representation methods; Application to image compression and indexing," *Hum.-Comput. Interact.*, 2010.

[39] R. Rubinstein, M. Zibulevsky, and M. Elad, "Double sparsity: Learning sparse dictionaries for sparse signal approximation," *IEEE Trans. Signal Process.*, vol. 58, no. 3, pp. 1553–1564, Mar. 2010.

[40] S. K. Sahoo and A. Makur, "Sparse sequential generalization of $K$-means for dictionary training on noisy signals," *Signal Process.*, vol. 129, pp. 62–66, 2016.

[41] I. Horev, O. Bryt, and R. Rubinstein, "Adaptive image compression using sparse dictionaries," in *Proc. IEEE Int. Conf. Syst., Signals Image Process. (IWSSIP)*, Apr. 2012, pp. 592–595.

[42] Y. Sun, M. Xu, X. Tao, and J. Lu, "Online dictionary learning based intra-frame video coding," *Wireless Pers. Commun.*, vol. 74, no. 4, pp. 1281–1295, 2014.

[43] O. Bryt and M. Elad, "Compression of facial images using the K-SVD algorithm," *J. Vis. Commun. Image Represent.*, vol. 19, no. 4, pp. 270–282, 2008.

[44] M. Elad, R. Goldenberg, and R. Kimmel, "Low bit-rate compression of facial images," *IEEE Trans. Image Process.*, vol. 16, no. 9, pp. 2379–2383, Sep. 2007.

[45] J. Zepeda, C. Guillemot, and E. Kijak, "Image compression using sparse representations and the iteration-tuned and aligned dictionary," *IEEE J. Sel. Topics Signal Process.*, vol. 5, no. 5, pp. 1061–1073, Sep. 2011.

[46] H. Yang, W. Lin, and C. Deng, "Learning based screen image compression," in *Proc. IEEE Int. Workshop Multimedia Signal Process. (MMSP)*, Sep. 2012, pp. 77–82.

[47] G. Shao, Y. Wu, Y. A. X. Liu, and T. Guo, "Fingerprint compression based on sparse representation," *IEEE Trans. Image Process.*, vol. 23, no. 2, pp. 489–501, Feb. 2014.

[48] M. Nejati, S. Samavi, N. Karimi, S. M. R. Soroushmehr, and K. Najarian, "Boosted dictionary learning for image compression," *IEEE Trans. Image Process.*, vol. 25, no. 10, pp. 4900–4915, Oct. 2016.

[49] K. Skretting and K. Engan, "Image compression using learned dictionaries by RLS-DLA and compared with K-SVD," in *Proc. IEEE Int. Conf. Acoust., Speech Signal Process. (ICASSP)*, May 2011, pp. 1517–1520.

[50] Y. Yuan, O. C. Au, A. Zheng, H. Yang, K. Tang, and W. Sun, "Image compression via sparse reconstruction," in *Proc. IEEE Int. Conf. Acoust., Speech Signal Process. (ICASSP)*, May 2014, pp. 2025–2029.

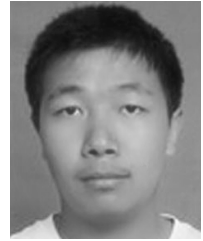

**Xiang Zhang** received the B.S. degree in computer science from the Harbin Institute of Technology, Harbin, China, in 2013. He is currently pursuing the Ph.D. degree. He was a Visiting Student with the Information Processing Laboratory, University of Washington, Seattle, WA, USA, in 2017. His research interests include video compression, image/video quality assessment, and visual retrieval.

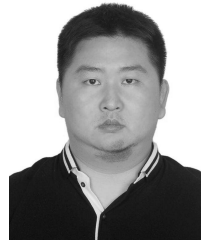

**Jiarui Sun** received the B.S. degree in mathematics and applied mathematics from Peking University, Beijing, China, in 2008, and the M.Sc. degree from Kyoto University, Kyoto, Japan, in 2013. He is currently pursuing the D.Eng. degree with Peking University, Shenzhen, China. He is an Algorithm Engineer with Shenzhen City Tencent Computer System Co. Ltd., where he was involved in artificial intelligence algorithm on medical images. His research interests and areas of publication include image processing, deep learning, and medical signal processing.

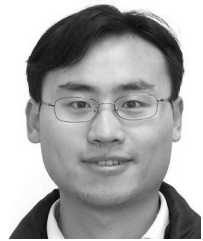

**Siwei Ma** (S'03–M'12) received the B.S. degree from Shandong Normal University, Jinan, China, in 1999, and the Ph.D. degree in computer science from the Institute of Computing Technology, Chinese Academy of Sciences, Beijing, China, in 2005. From 2005 to 2007, he held a post-doctoral position at the University of Southern California, Los Angeles, USA. Then, he joined the Institute of Digital Media, School of Electronics Engineering and Computer Science, Peking University, Beijing, where he is currently a Professor. He has published over 100 technical articles in refereed journals and proceedings in the areas of image and video coding, video processing, video streaming, and transmission.

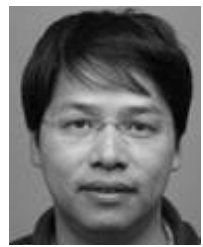

**Zhouchen Lin** (M'00–SM'08–F'18) received the Ph.D. degree in applied mathematics from Peking University in 2000. He is currently a Professor with the Key Laboratory of Machine Perception, School of Electronics Engineering and Computer Science, Peking University. His research interests include computer vision, image processing, machine learning, pattern recognition, and numerical optimization. He is a Senior Program Committee Member of AAAI 2016–2018 and IJCAI 2016/2018. He is a fellow of the IAPR. He is an Area Chair of CVPR 2014/2016, ICCV in 2015, and NIPS in 2015. He is an Associate Editor of the IEEE TRANSACTIONS ON PATTERN ANALYSIS AND MACHINE INTELLIGENCE and the INTERNATIONAL JOURNAL OF COMPUTER VISION.

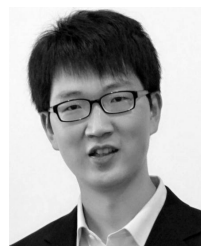

**Jian Zhang** (M'14) received the B.Sc. degree from the Department of Mathematics, Harbin Institute of Technology (HIT), Harbin, China, in 2007, and the M.Eng. and Ph.D. degrees from the School of Computer Science and Technology, HIT, in 2009 and 2014, respectively. He is currently an Assistant Professor with the School of Electronic and Computer Engineering, Peking University Shenzhen Graduate School. His research interests include image/video restoration and compression, optimization, and deep learning. He was a recipient of the Best Paper Award and the Best Student Paper Award at the IEEE VCIP in 2011 and 2015, respectively.

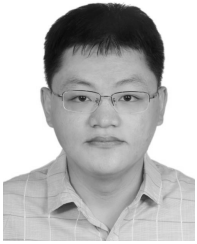

**Shiqi Wang** (M'15) received the B.S. degree in computer science from the Harbin Institute of Technology in 2008 and the Ph.D. degree in computer application technology from Peking University in 2014. From 2014 to 2016, he was a Post-Doctoral Fellow with the Department of Electrical and Computer Engineering, University of Waterloo, Waterloo, Canada. From 2016 to 2017, he was with the Rapid-Rich Object Search Laboratory, Nanyang Technological University, Singapore, as a Research Fellow. He is currently an Assistant Professor with the Department of Computer Science, City University of Hong Kong. He has proposed over 30 technical proposals to ISO/MPEG, ITU-T, and AVS standards. His research interests include video compression, processing, and analysis.

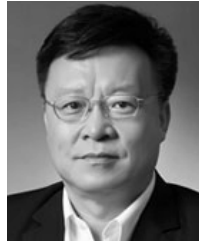

**Wen Gao** (M'92–SM'05–F'09) received the Ph.D. degree in electronics engineering from The University of Tokyo, Japan, in 1991. He was a Professor of computer science with the Harbin Institute of Technology, from 1991 to 1995, and a Professor with the Institute of Computing Technology, Chinese Academy of Sciences. He is currently a Professor of computer science with Peking University, China.. He has published extensively including five books and over 600 technical articles in refereed journals and conference proceedings in the areas of image processing, video coding and communication, pattern recognition, multimedia information retrieval, multimodal interface, and bioinformatics. He chaired a number of prestigious international conferences on multimedia and video signal processing, such as the IEEE ICME and the ACM Multimedia, and also served on the advisory and technical committees of numerous professional organizations. He served or serves on the Editorial Board for several journals, such as the IEEE TRANSACTIONS ON CIRCUITS AND SYSTEMS FOR VIDEO TECHNOLOGY, the IEEE TRANSACTIONS ON MULTIMEDIA, the IEEE TRANSACTIONS ON IMAGE PROCESSING, the IEEE TRANSACTIONS ON AUTONOMOUS MENTAL DEVELOPMENT, the *EURASIP Journal of Image Communications*, and the *Journal of Visual Communication and Image Representation*.